\documentclass{article}

\usepackage{PRIMEarxiv}

\usepackage[utf8]{inputenc} 
\usepackage[T1]{fontenc}    
\usepackage{hyperref}       
\usepackage{url}            
\usepackage{booktabs}       
\usepackage{amsfonts}       
\usepackage{nicefrac}       
\usepackage{microtype}      
\usepackage{lipsum}
\usepackage{fancyhdr}       
\usepackage{graphicx}       
\graphicspath{{media/}}     

\usepackage{booktabs}
\usepackage{caption}
\usepackage{subcaption}
\usepackage{amsmath}

\title{Statistical and Spatio-temporal Hand Gesture Features for Sign Language Recognition using the Leap Motion Sensor
}

\author{
  Jordan J. Bird \\
  Computational Intelligence and Applications Research Group (CIA) \\
  Nottingham Trent University \\
  Nottingham, UK\\
  \texttt{jordan.bird@ntu.ac.uk} \\
}

\begin{document}
\maketitle

\begin{abstract}
In modern society, people should not be identified based on their disability, rather, it is environments that can disable people with impairments. Improvements to automatic Sign Language Recognition (SLR) will lead to more enabling environments via digital technology. Many state-of-the-art approaches to SLR focus on the classification of static hand gestures, but communication is a temporal activity, which is reflected by many of the dynamic gestures present. Given this, temporal information during the delivery of a gesture is not often considered within SLR. The experiments in this work consider the problem of SL gesture recognition regarding how dynamic gestures change during their delivery, and this study aims to explore how single types of features as well as mixed features affect the classification ability of a machine learning model. 18 common gestures recorded via a Leap Motion Controller sensor provide a complex classification problem. Two sets of features are extracted from a 0.6 second time window, statistical descriptors and spatio-temporal attributes. Features from each set are compared by their ANOVA F-Scores and p-values, arranged into bins grown by 10 features per step to a limit of the 250 highest-ranked features. Results show that the best statistical model selected 240 features and scored 85.96\% accuracy, the best spatio-temporal model selected 230 features and scored 80.98\%, and the best mixed-feature model selected 240 features from each set leading to a classification accuracy of 86.75\%. When all three sets of results are compared (146 individual machine learning models), the overall distribution shows that the minimum results are increased when inputs are any number of mixed features compared to any number of either of the two single sets of features.
\end{abstract}


\section{Introduction}
One of the most important subfields of human activity recognition is the applied intelligence for sign language recognition, wherein the goal is to provide systems which can translate sign language to written text via the classification of individual gestures which pertain to said words and phrases. The ability to communicate is often taken for granted, and in the deaf community, a lack of communication leads to isolation and depression\cite{charlson1992successful,sheppard2010lived}. Studies have shown that isolation can be reduced through computer-mediated communication\cite{bishop2000computer}, that is, the use of computational techniques to provide a model-in-the-middle approach for bridging a communicative barrier between those who can and cannot use sign language to an effective enough level. The aforementioned 1992 study noted that teenagers often experience this when attempting to communicate with their parents and at school, and it has also been observed that members of the elderly community who are deaf experience isolation when entering a nursing home designed for hearing residents\cite{greene1997residential}. 

According to the World Health Organisation\cite{WHO_2021}, more than 1.5 billion people are affected by some form of hearing loss. Of those, 430 million are considered to have hearing loss which is disabling in modern society. It is also noted that this is a growing problem, the number of people projected to have hearing loss by 2050 is 2.5 billion, with 700 million of those forecast to be considered to have disabling hearing loss. These figures argue for the development of improved methods for sign language communication, given that few education systems implement this form of communication in their curricula. This article explores how different types of features can be extracted from hand gestures to improve the classification of said gesture, that is, translating a physical gesture to an on-screen text. Such a technology would allow for better communication between those who can and cannot communicate through physical gestures. Literature review notes that, unlike speech recognition which is commercially viable, automatic Sign Language Recognition remains in its infancy. Many problems are encountered, one of which is the processing of only spatial observations for static gestures. Studies which perform feature extraction beyond the information provided by sensor APIs tend to show improved results and reduced volatility, and so this study will explore how features of differing natures can be used to classify hand gestures along with how they can be fused to compliment one another. Many sign language gestures are dynamic and occur over more than one point in time, and many studies do not consider this temporal attribute when processing hand gesture data. It is thus a goal of this study to also explore how the implementation of spatio-temporal features can improve the overall classification of dynamic gestures. 

The main scientific contributions of this work are as follows:
\begin{enumerate}
    \item The extraction of statistical and spatio-temporal features from a dataset of 18 different gestures.
    \item The analysis of the extracted gestures by their ANOVA F-scores and ranking, as well as their p-values.
    \item The training and analysis of machine learning models when considering either one or both sets of features differing in number, leading to a total of 146 models trained.
    \item Comparison of results shows that hand gesture recognition is improved when a mixed set of features are considered, leading to an overall mean classification accuracy of 86.75\% (240 statistical and 240 spatio-temporal features).
\end{enumerate}

\section{Background and Related Work}
Sign Language Recognition is the study of how algorithms can be engineered to automate the translation and transcription of physical, facial, and hand gestures to written text. As noted in \cite{cooper2011sign}, automatic speech recognition is advanced enough for commercial viability, but automatic Sign Language Recognition (SLR) is still an early technology. Although this book was written in 2011, SLR is still yet to be commercially viable in society, and further work is needed to improve the technology. Wadhawan and Kumar's 2021 literature review on a decade of SLR research\cite{wadhawan2021sign} noted the growing trend of published works, which doubled between 2013 and 2017. The review also notes that much of the research has been performed for static gestures, which are nontemporal and thus easier to classify than dynamic gestures which may convey a single word or feeling in their entirety. It is for this reason that the experiments presented in this work attempt to classify dynamic features by considering their statistical and temporal behaviours observed within a time-window.

Several sensors have been proposed as candidates for automatic Sign Language Recognition, the most prominent being RGB cameras \cite{imagawa2000recognition,parelli2020exploiting}, depth-sensing cameras \cite{kuznetsova2013real,agarwal2013sign,aly2019user}, smart gloves\cite{mehdi2002sign,kuroda2004consumer,oz2011american}, and biological signal processing of electroencephalography\cite{alqattan2017towards} and electromyography\cite{kosmidou2006evaluation,savur2015real}. This literature review mainly focuses on the Leap Motion Controller sensor, which is used in this study. The Leap Motion Controller operates in infrared with an array of two cameras to discern where in space the hands are located. The sensor's API allows for the measurement of basic spatial features along with velocity of some points of the hands and arms. In \cite{chuan2014american}, the authors proposed using KNN and SVM models to classify American Sign Language alphabet gestures via a Leap Motion Controller. Through 4-fold cross validation, KNN had a mean accuracy of 72.87\% which was outperformed by a Radial Basis Function SVM at 79.83\%. Features were smoothed via a sliding window technique to improve classification ability. Similarly, in \cite{mohandes2014arabic}, the authors proposed Bayesian and Deep Learning approaches to leap motion-based Arabic Sign Language recognition trained via 5-fold cross validation; a Naive Bayes classifier scored around 98\% and deep neural networks experienced 99\% classification accuracy. In the aforementioned work, the authors chose to select half of the features provided by the Leap Motion API which were most relevant. In addition, feature extraction was performed to further extract the mean values from each frame for the relevant features. The results in this study also show the improvement when such features are engineered, suggesting that further extractions from those provided by the sensor's software provide a set of attributes which are useful for the task. Temporal learning was noted to be useful in Indian Sign Language recognition in \cite{mittal2019modified}, where Long Short Term Memory models could classify 35 individual gestures with 89.5\% accuracy which led to 72.3\% sentence accuracy. The authors in the Indian Sign Language study noted that LSTMs with a depth of three layers were the most apt to extract temporal features for classification. The experiments performed in \cite{vaitkevivcius2019recognition} focused on the classification of the American Sign Language alphabet via leap motion data after recording 18 different pangrams, noting the effectiveness of the Hidden Markov Model for classification, which achieved a mean accuracy of 86.1\%. The choice of model is especially interesting, similarly to the LSTM study, where sequential (temporal) observations enable better gesture recognition. Geometric Template Matching was proposed as an effective model for the recognition of the American Sign Language alphabet within \cite{khan2016sign}, which achieved around 52.56\% accuracy; the authors noted that letters A, B, D, and I were classified perfectly whereas issues were encountered when letters P, R, and T could not be classified correctly by the model. Multi-modality is also considered as a candidate for improving the state-of-the-art in automatic SLR; the authors in \cite{bird2020british} proposed the late fusion of image and leap motion attributes for British and American Sign Language recognition, achieving 94.44\% and 82.55\% accuracy metrics on the datasets, respectively. The aforementioned previous study achieved 72.73\% accuracy on the leap motion data alone, which is the same dataset used in the experiments in this article. Zhang et al.'s 2019 study\cite{zhang2019multimodal} discovered that multi-modality could drastically improve sign recognition when fusing RGB and Depth data. The model presented by the study was computationally expensive, requiring two VGG16 convolutional neural networks to process the sensor information. Similar findings came from Gao et al.\cite{gao2019two} based on a dual-CNN approach where image enhancement and pixel mapping were fused. The main difference between the aforementioned two studies is that Zhang et al. proposed fusion of extracted features through a tertiary neural network, whereas Gao et al. fused the predictions of two separate models by treating the softmax activation vectors as features for a tertiary classifier. The framework presented by \cite{kelly2009framework} proposed the fusion of hand gesture and non-manual (facial expressions and non-hand movements) through Hidden Markov Models, again achieving a better result when further data was considered prior to predictions. 

With the literature review in mind, it seems that three of the prime candidates for the improvement of sign language recognition are feature extraction, the consideration of temporal events, and multi-modality. It is for this reason that this study focuses on the differences between statistical descriptors and spatio-temporal information as mixed multi-modal inputs to a learning algorithm. The main research questions to explore here surround how the two sets of features perform regarding gesture recognition, and whether mixing the attributes lead to a better result overall. 

\section{Method}
This section explains the methodology followed by the experiments in this work. This includes data collection, feature extraction and analysis, and machine learning approaches for the collection of results prior to comparison.
\subsection{Data collection}
Data collection is performed from a previous study which fused hand gesture and photographic data for the classification of BSL\cite{bird2020british}. From this dataset, only the hand gesture data collected from a Leap Motion sensor is used. The following gestures present as the 18-class problem: \textit{Hello/Goodbye, You/Yourself, Me/Myself, Name, Sorry, Good, Bad, Excuse Me, Thanks/Thank you, Time, Airport, Bus, Car, Aeroplane, Taxi, Restaurant, Drink, and Food}. These gestures were selected due to their usefulness in communication. For each of the gestures, 3D data was recorded from the leap motion sensor in the form of:
\begin{itemize}
    \item \textbf{Arms:} Start position of the arm (X, Y, and Z), end position of the arm (X, Y, and Z), 3D angle between the start and end positions of the arm, and velocity of the arm (X, Y, and Z)
    \item \textbf{Elbows:} Position of the elbow (X, Y, and Z).
    \item \textbf{Wrists:} Position of the wrist (X, Y, and Z).
    \item \textbf{Palms:} Pitch, Yaw, Roll, 3D angle of the palm, position of the palm (X, Y, and Z), velocity of the palm (X, Y, and Z), and normal of the palm (X, Y, and Z).
    \item \textbf{Fingers:} Direction of the finger (X, Y, and Z), position of the finger (X, Y, and Z), and velocity of the finger (X, Y, and Z).
    \item Finger joints: Start position of the joint (X, Y, and Z), end position of the joint (X, Y, and Z), 3D angle of the joint, direction of the finger (X, Y, and Z), position of the joint (X, Y, and Z), and velocity of the joint (X, Y, and Z).
\end{itemize}
3D angles ($\theta$) are computed via the following:
\begin{equation}
     \theta  = arccos \left( \frac{ab}{  \mid a\mid \mid b \mid } \right),
\end{equation}
where $\mid a\mid$ and $\mid b\mid$ are:
\begin{equation}
\begin{split}
    \mid a\mid =  \sqrt{ {a_{x}}^2 + {a_{y}}^2 + {a_{z}}^2 } \\
    \mid b\mid =  \sqrt{ {b_{x}}^2 + {b_{y}}^2 + {b_{z}}^2 } ,
\end{split}
\end{equation}
regarding the $x, y$, and $z$ co-ordinates of each hand/arm point recorded. Thus, the dataset presents static data (locations in space) as well as a limited amount of temporal information (velocity of joints). Otherwise, the dataset does not describe motion over short periods of time, which is useful for the understanding of gestures. It is for this reason that this study explores how different types of features affect classification ability, detailed further in Section \ref{subsec:featureex}. 

\subsection{Feature Extraction and Learning}
\label{subsec:featureex}
The dataset was recorded at 5Hz, i.e., once every 0.2 seconds. Time windows of 0.6 (three vectors) are used in this study for two reasons: (i) shorter times cause issues with extracting a number of features, and (ii) time windows longer than 3 would cause communication to become unnatural and slow. Two sets of features are extracted within this study, statistical and spatio-temporal. The statistical features extracted from each point were:
\begin{enumerate}
    \item Absolute energy
    \item Average power
    \item Empirical Cumulative Distribution Function (ECDF)
    \item ECDF percentiles (0.2 and 0.8)
    \item Cumulative sum of samples less than ECDF percentiles (0.2 and 0.8)
    \item Shannon Entropy
    \item Histogram of the data (10 bins)
    \item Interquartile range
    \item Kurtosis
    \item Maximum value
    \item Mean value
    \item Mean absolute deviation
    \item Median value
    \item Median absolute deviation
    \item Minimum value
    \item Peak to peak distance
    \item Root mean square
    \item Skewness
    \item Standard deviation
    \item Variance
\end{enumerate}
The spatio-temporal features extracted from each point were:
\begin{enumerate}
    \item Area under curve (computed via the trapezoid rule)
    \item Autocorrelation
    \item Centroid along the time axis
    \item Mean differences
    \item Mean absolute differences
    \item Median differences
    \item Median absolute differences
    \item Positive turning points
    \item Negative turning points
    \item Neighbourhood peaks (10)
    \item Slope via fitting a linear equation
    \item Sum of absolute differences
    \item Zero crossing rate
\end{enumerate}

Given that many features are extracted, and it is unknown which are useful, analysis of variance (ANOVA) testing is performed on each of the features to rank them. Those with a p-value greater than 0.05 are discarded, and then the top 250 features are taken from each set to form datasets for classification. Classification is performed on $\{10, 20, 30, ..., 250\}$ input features (ranked from best to worst) for each set to form the singular set models. For mixing features, two methods are used in order to allow for all features to be present; firstly, a total of 250 features by using $\{10, 20, 30, ..., 240\}$ and $\{240, 230, 220 ..., 10\}$ from each of the sets, and then selecting $\{10, 20, 30, ..., 250\}$ from both sets of features. This leads to a total of 146 individual machine learning models to compare based on their input type(s). The classifier selected for this experiment is a Random Forest of 100 estimators given its nature not to overfit to training data, future work notes the possibility of exploring other models based on the findings within this study. 

Feature selection and model training are implemented via scikit-learn\cite{scikit-learn}, and feature extraction is implemented via the TSFEL library\cite{liu2020tsfel}. Random states for all experiments are set to 1 for comparability of each approach, where random numbers are generated by an Intel Core i7-8700K, Python 3.7.9, and scikit-learn 1.0.2.

\section{Results and Discussion}
In this section, the analysis of features and the experimental results are presented, discussed, and compared. This includes the preprocessing of raw data, the extraction and analysis of both statistical and spatio-temporal features, classification of the two sets of features, fusion of the feature sets, and finally an overall comparison and analysis of all results collected throughout the experiments. 

\subsection{Raw Data Pre-processing}
Since extracting all available features from all recorded hand gesture features would result in exceedingly large datasets leading to an equally large number of experiments with resource-intensive requirements, feature selection is performed and analysed to provide an initial set of features for statistical and temporal extraction.

\begin{figure}
    \centering
    \includegraphics[scale=0.8]{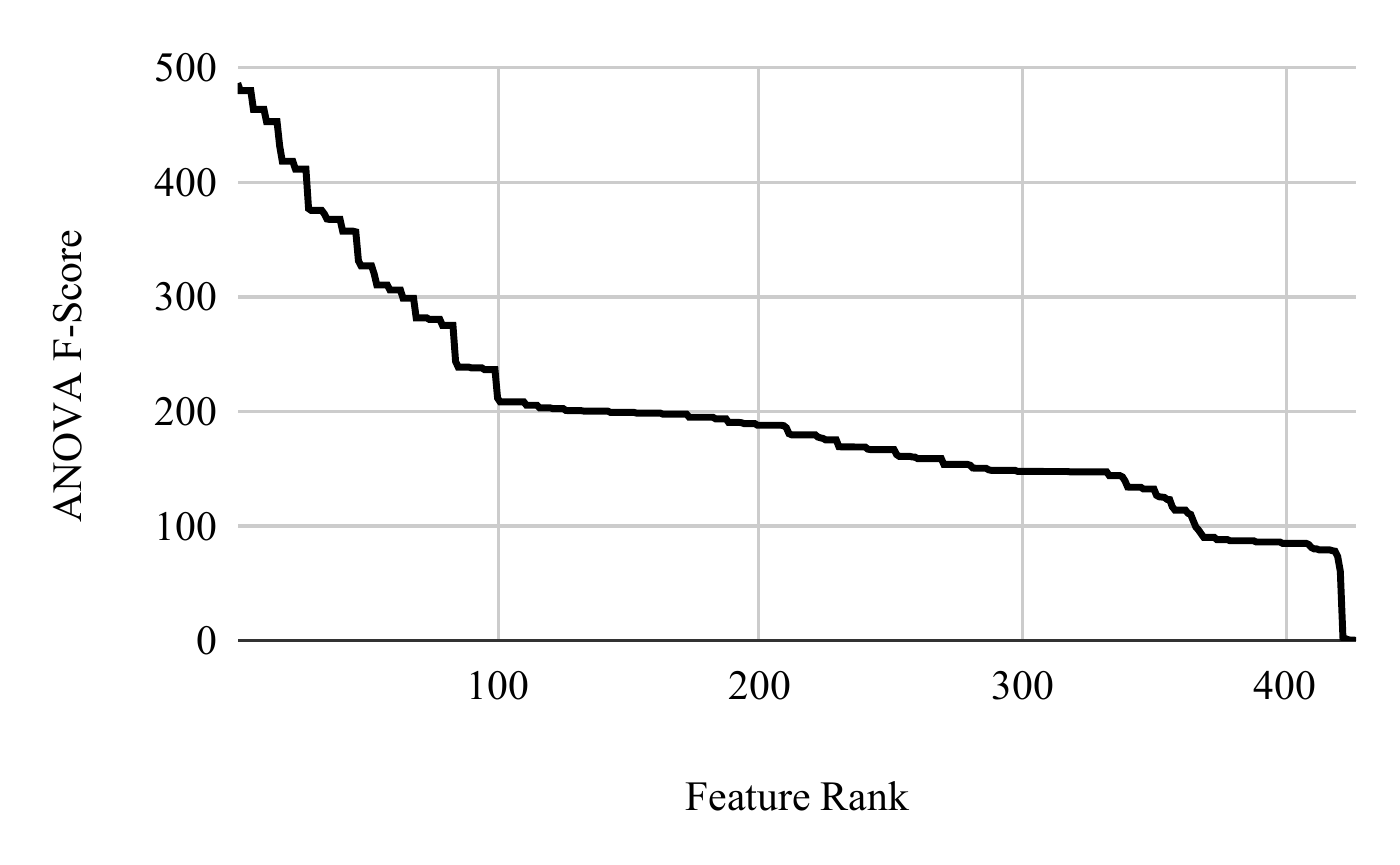}
    \caption{ANOVA F-Score for each of the raw hand features, ordered by highest to lowest.}
    \label{fig:anova-fscores}
\end{figure}

\begin{figure}
    \centering
    \includegraphics[scale=0.8]{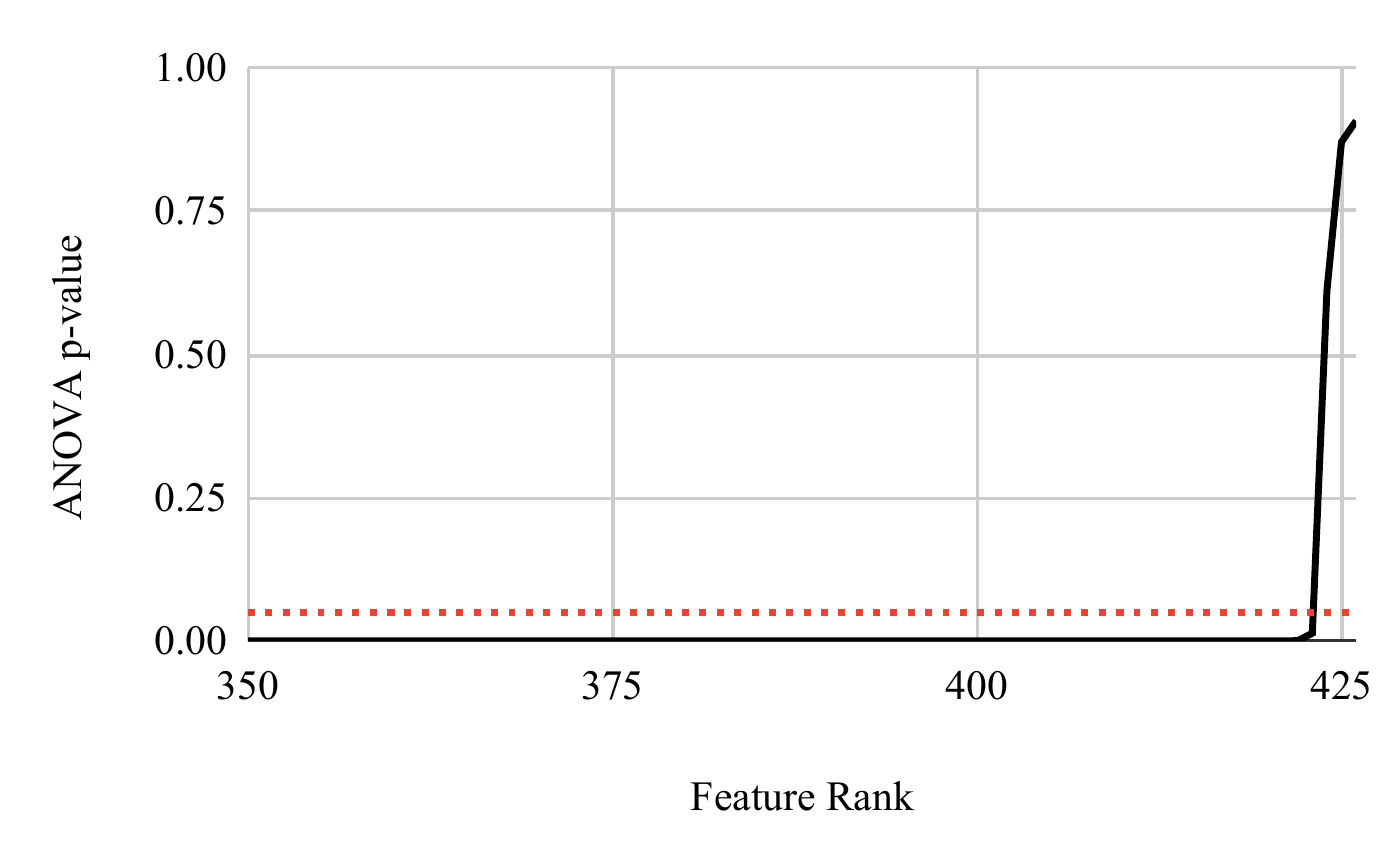}
    \caption{ANOVA p-values for each of the raw hand features, ordered by ANOVA F-Score ranking. Dashed line shows $p=0.05$.}
    \label{fig:anova-pvalues}
\end{figure}

Figure \ref{fig:anova-fscores} shows the F-Scores for each of the features by ranking, the first 99 features show relatively high scores compared to the remainder of the data. A noticeable dropoff point exists for several of the features ranked bottom towards the end of the graph, noting that they are exceptionally useless for classification as opposed to the prior dataset. Figure \ref{fig:anova-pvalues} show the p-values for each of these features ranked in the same order, note that the statistically insignificant values correlate with those which had the lowest ANOVA F-Scores. There were four features in particular where $p>0.05$, in order of smallest to largest they were; the direction of the left hand on the y-axis ($p=0.61$), the velocity of the left palm on the y-axis ($p=0.87$), the direction of the left hand on the z-axis ($p=0.9$), and finally the velocity of the right palm on the x-axis ($p=0.98$). 
\begin{table}[]
\centering
\caption{Mean ANOVA F-Scores for the top N-ranked features}
\label{tab:anova-average}
\begin{tabular}{@{}ll@{}}
\toprule
\textbf{Top-N Features} & \textbf{Mean ANOVA F-Score} \\ \midrule
1                       & 486.32               \\
50                      & 405.12             \\
100                     & 339.6            \\
150                     & 294.25             \\
200                     & 269.47            \\
250                     & 250.73            \\
300                     & 234.66            \\
350                     & 221.66            \\
400                     & 206            \\
427 (all)                     & 197.7            \\ \bottomrule
\end{tabular}
\end{table}
Table \ref{tab:anova-average} shows the average ANOVA F-Score for sets of features, with the top-ranked features through groups increasing by 50. The two initial drop-off points can be seen to affect the value when 100 features are considered. The top 50 features are used as the set for feature extraction in all the following experiments, future work notes the exploration of the size of this set of features as being something to explore based on the results discovered by this work.  

\subsection{Extraction and Analysis of Statistical and Spatio-temporal Features}
\begin{figure}
    \centering
    \includegraphics[scale=0.8]{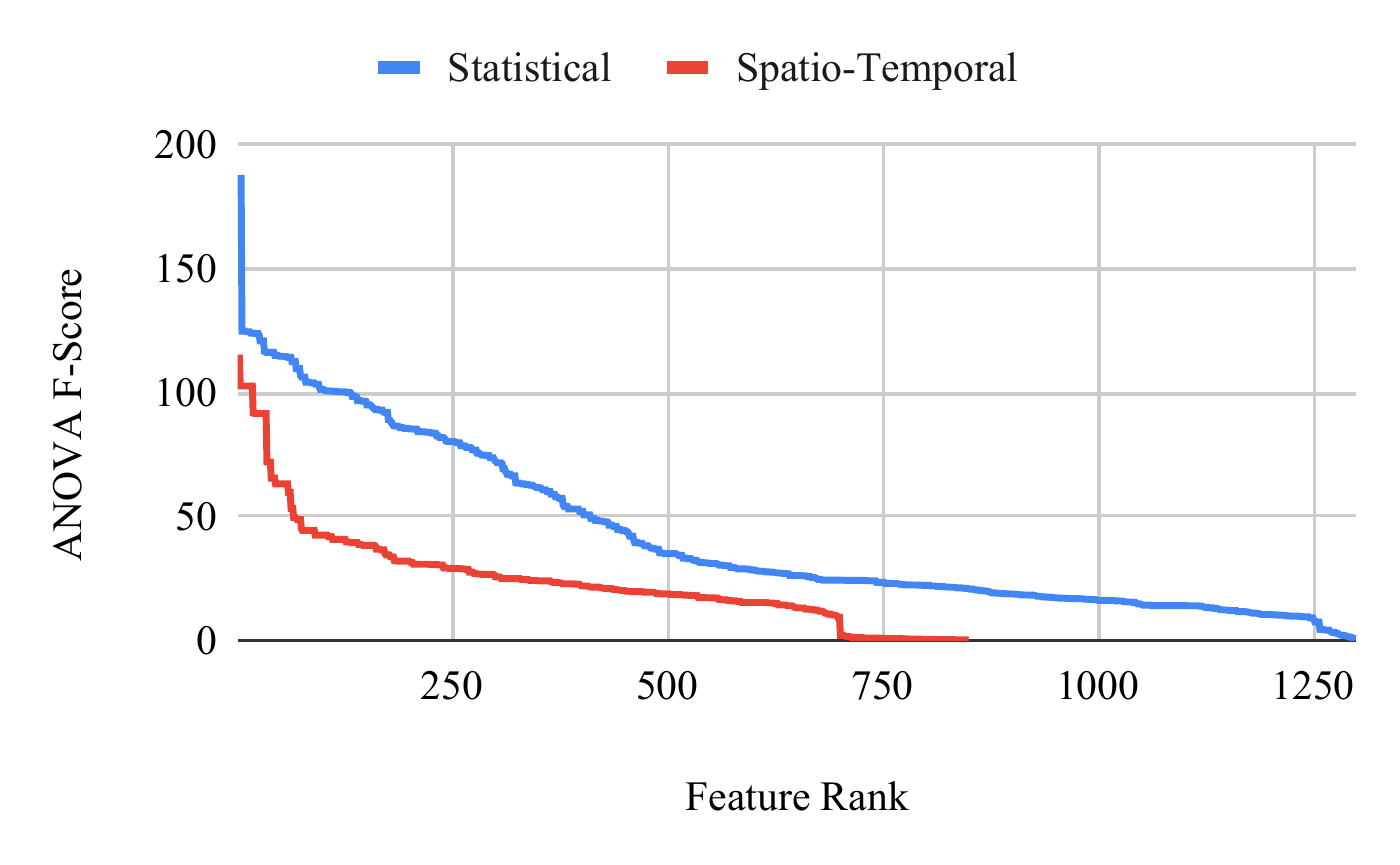}
    \caption{ANOVA F-Score for the extracted statistical and spatio-temporal features.}
    \label{fig:anova-fscore-features}
\end{figure}

In this section, features are extracted and analysed via feature scoring techniques. Figure \ref{fig:anova-fscore-features} shows the ANOVA F-scores for each of the extracted features ordered by score. It can be observed that the best 64 statistical features are considered to score higher than all spatio-temporal features. It can also be observed that there are a greater number of statistical features considered useful for classification than there are useful spatio-temporal features. 
\begin{figure}
    \centering
    \includegraphics[scale=0.8]{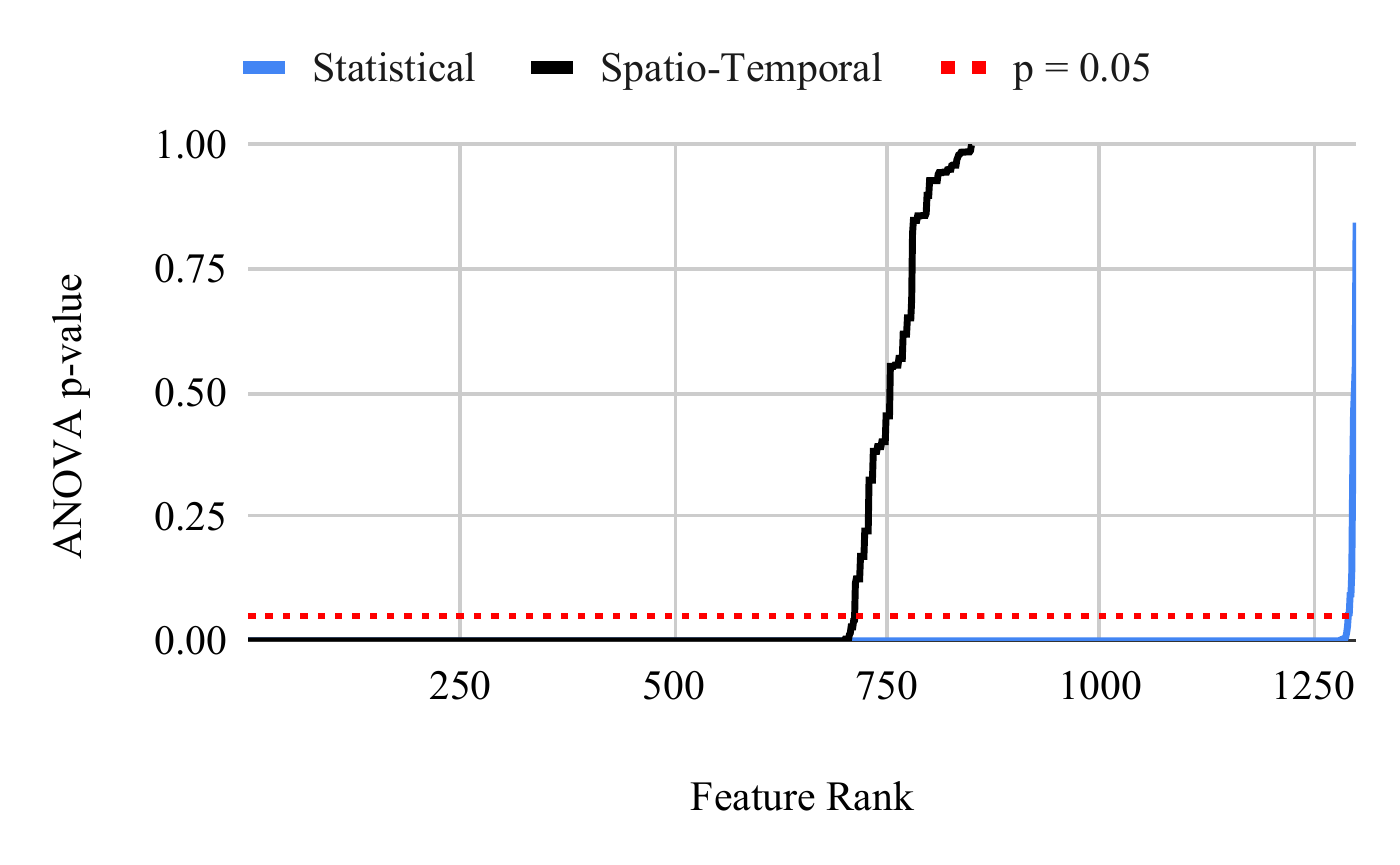}
    \caption{ANOVA F-Score p-values for the extracted statistical and spatio-temporal features.}
    \label{fig:anova-pvalues-features}
\end{figure}
Four statistical features were ranked first with $F=186.54$. These were the fourth Empirical Cumulative Distribution of the distal end of the right index finger on the z-axis, the fourth ECDF of the distal end of the right thumb on the z-axis, the fourth ECDF of the right ring finger on the z-axis, and the fourth ECDF of the right middle finger on the z-axis. Following these were the percentile count measurements of these same features. Ten statistical features had $P>0.05$, the highest being the second histogram of the pitch of the right hand where $p=0.84$ and $F=0.66$. The highest p-value statistical feature where $p\leq0.05$ was the eighth histogram of the right palm's velocity on the z-axis where $p=0.0295$ and $F=1.74$. 
In terms of spatio-temporal features, three were ranked first with $F=114.04$. These were the area under the curve for the right hand pitch, the zero crossing rate for the right hand's direction on the x-axis, and the sum of the absolute difference of the right hand pitch. 137 of the worst features in this set had $p>0.05$.

\begin{table}[]
\centering
\caption{Classification based on predictions by the most common class compared to a single attribute (computed by error rate).}
\label{tab:onerule}
\begin{tabular}{@{}llll@{}}
\toprule
\textbf{Domain}                   & \textbf{Selected Attribute}                                                   & \textbf{Correct} & \textbf{Accuracy} \\ \midrule
\textit{\textbf{N/A}}             & \begin{tabular}[c]{@{}l@{}}Most Common Class:\\ "GOOD"\end{tabular}           & 311/3291         & 9.45\%            \\
\textit{\textbf{Statistical}}     & \begin{tabular}[c]{@{}l@{}}ECDF Percentile 1:\\ Right Hand Pitch\end{tabular} & 885/3291         & 26.89\%           \\
\textit{\textbf{Spatio-temporal}} & \begin{tabular}[c]{@{}l@{}}Absolute Energy:\\ Right Hand Pitch\end{tabular}   & 826/3291         & 25.1\%            \\ \bottomrule
\end{tabular}
\end{table}

Although many statistical features seem more useful than those which are spatio-temporal, there exist useful features in both sets of data. Thus, classifiers may be improved by combining the two sets of features for multi-domain classification. Based on this, Table \ref{tab:onerule} shows a comparison of the lowest error rate for a single rule from both of the sets and classification by the most common class. By classifying based on the most common class, an accuracy of only 9.45\% is achieved, whereas considering a classification by a single attribute leads to accuracies of 26.89\% for the statistical attribute with the lowest error rate and 25.1\% for the spatio-temporal attribute with the lowest error rate. Although there is an observed disparity in ANOVA F-Scores, the classification abilities of the single best attributes from the two datasets are similar.

\subsection{Classification of Statistical Features}
In this section, the results are presented for the classification of sign language gesture using only the statistical features which were extracted.

\begin{figure}
    \centering
    \includegraphics[scale=0.8]{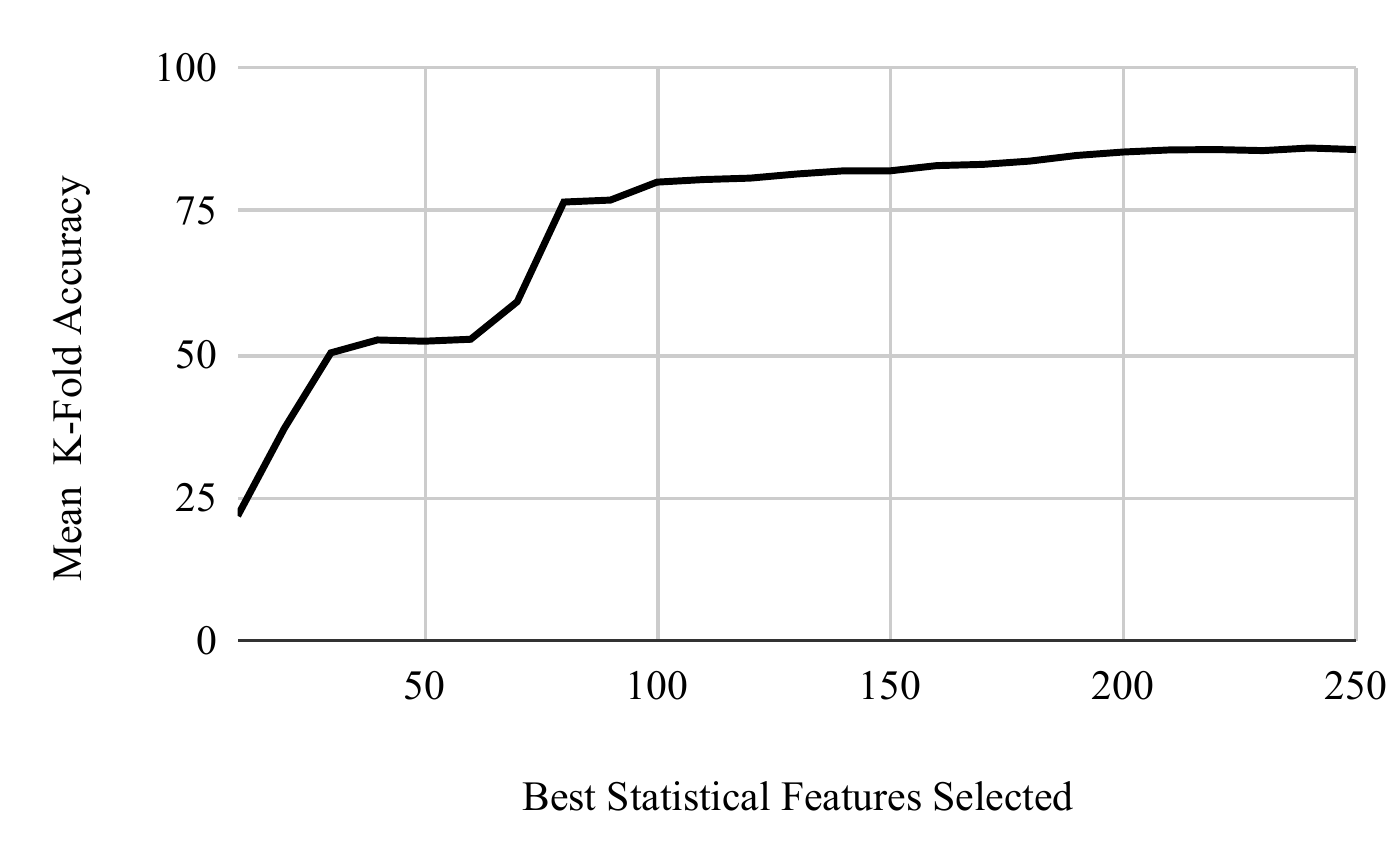}
    \caption{Mean K-Fold classification accuracy when selecting a growing number of the best statistical features.}
    \label{fig:rf-acc-statistical}
\end{figure}
\begin{figure}
    \centering
    \includegraphics[scale=0.8]{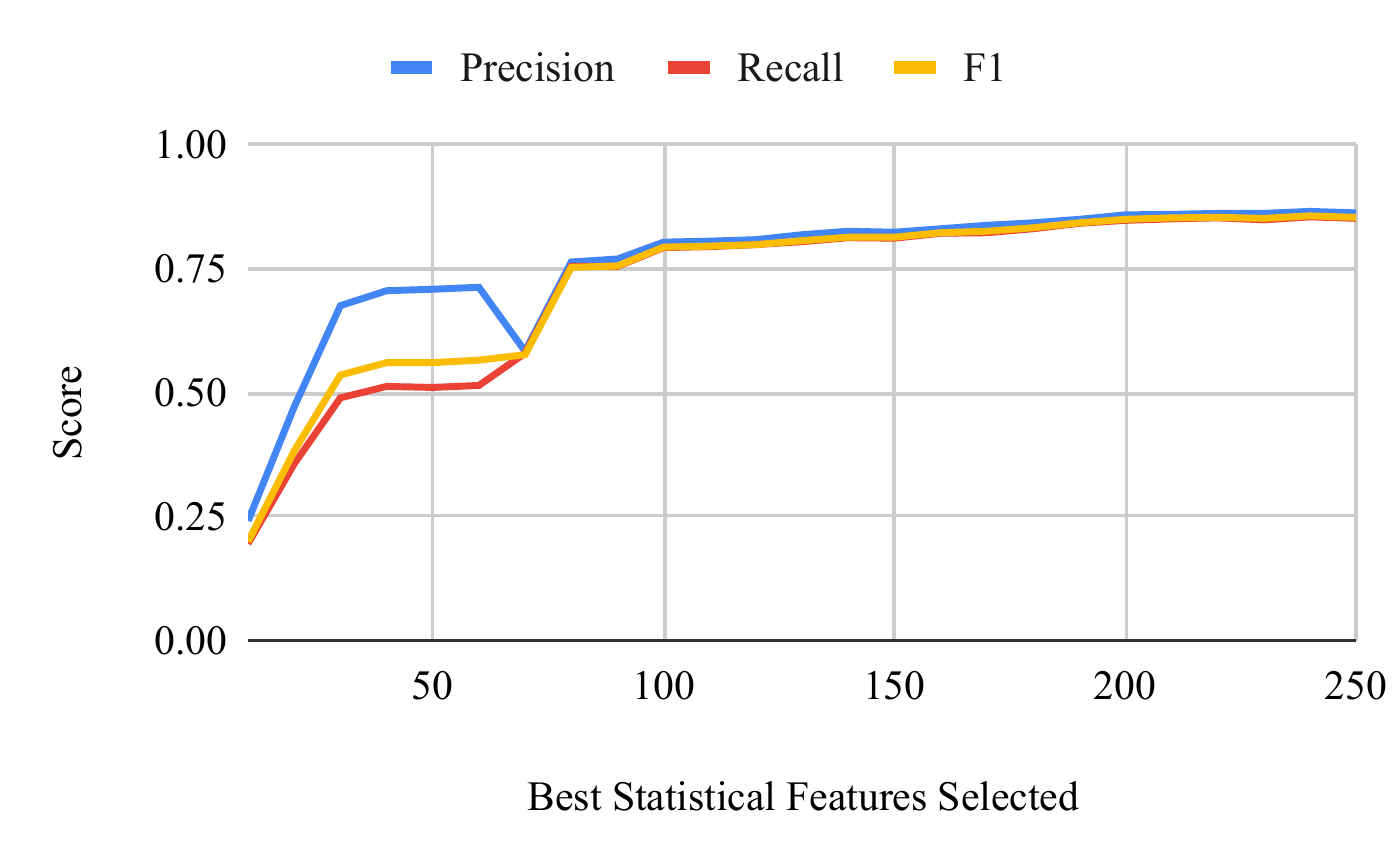}
    \caption{Mean classification metrics when selecting a growing number of the best statistical features.}
    \label{fig:rf-metrics-statistical}
\end{figure}

Figures \ref{fig:rf-acc-statistical} and \ref{fig:rf-metrics-statistical} show the accuracy and metrics of the Random Forest model when increasing the selected number of features. It can be observed that a selection of the best results causes the metrics to increase only slightly before a large increase at around 80 features. Following the 80 selected features, a slow increase is observed. The final selection of 250 features scored a mean 85.69\% accuracy which was the same as 220 features. These were second best when 240 features were selected, in which case the mean classification accuracy was 85.96\%. 240 features also led to the highest precision of 0.866, the highest recall of 0.855, and the highest F1-Score of 0.857.

\subsection{Classification of Spatio-temporal Features}
This subsection details the classification results when the set of extracted spatio-temporal features are presented as model training data. 

\begin{figure}
    \centering
    \includegraphics[scale=0.8]{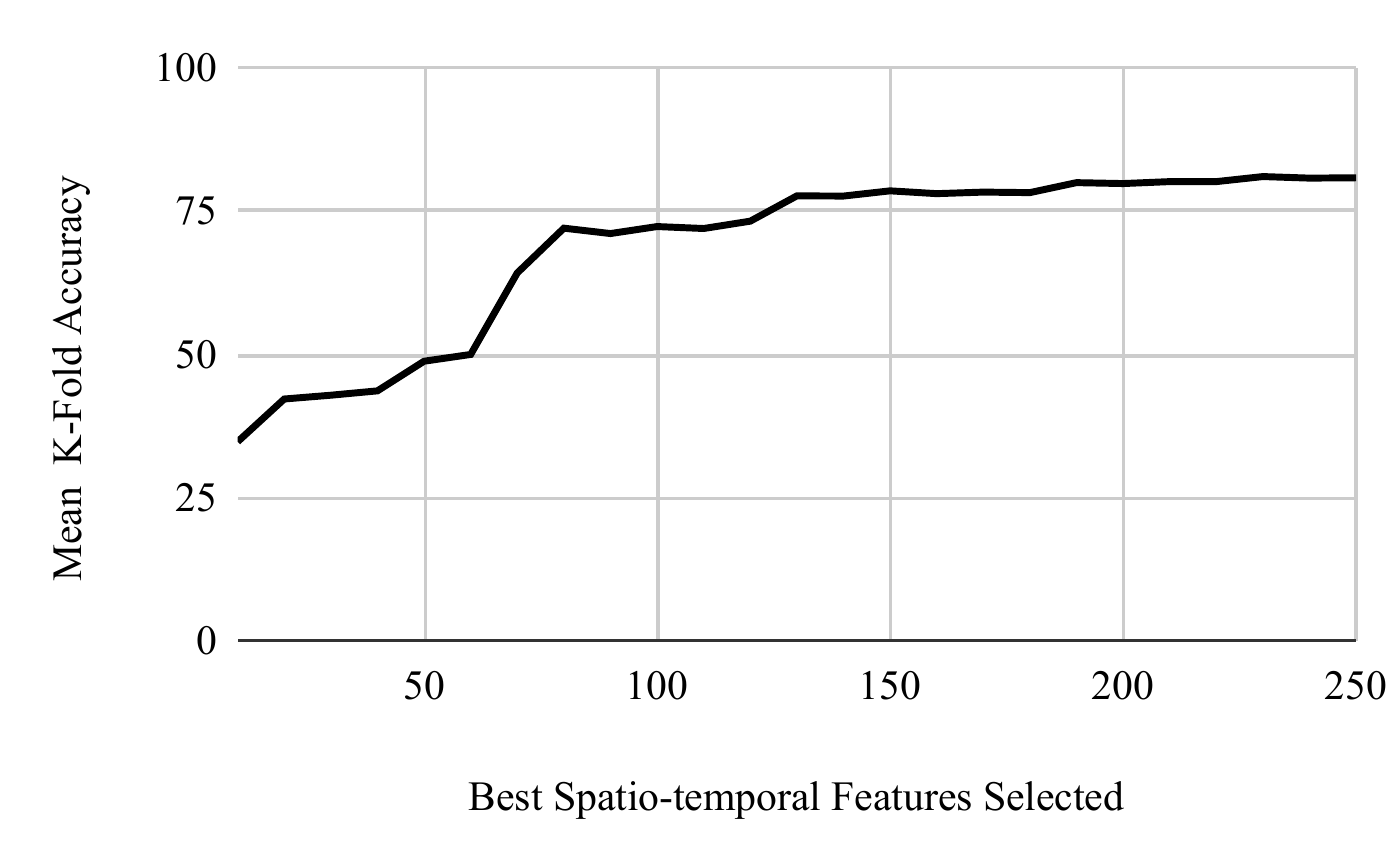}
    \caption{Mean K-Fold classification accuracy when selecting a growing number of the best spatio-temporal features.}
    \label{fig:rf-acc-temporal}
\end{figure}

\begin{figure}
    \centering
    \includegraphics[scale=0.8]{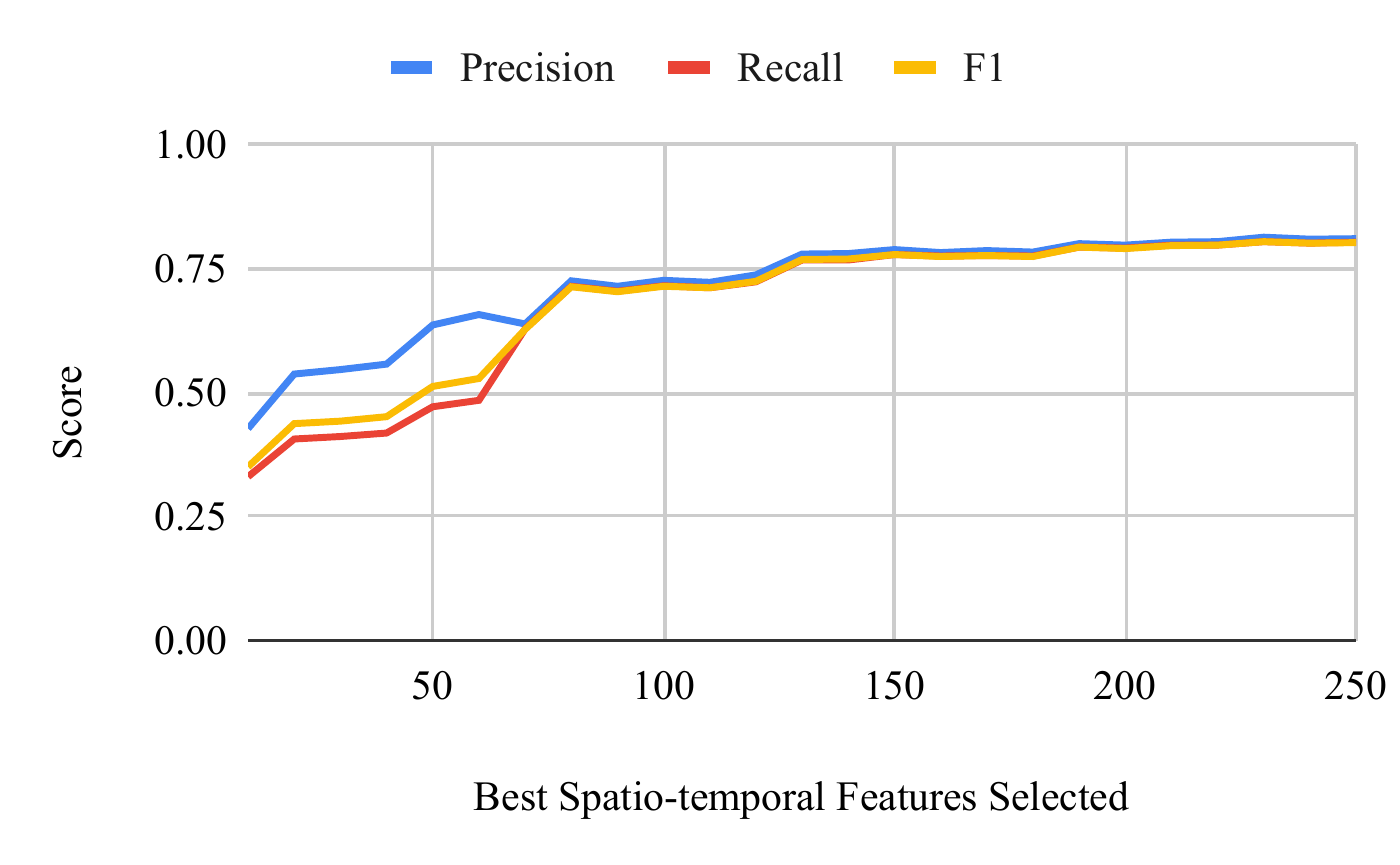}
    \caption{Mean classification metrics when selecting a growing number of the best spatio-temporal features.}
    \label{fig:rf-metrics-temporal}
\end{figure}

Similarly to the results presented in the previous section, Figures \ref{fig:rf-acc-temporal} and \ref{fig:rf-metrics-temporal} show the classification metrics when introducing more spatio-temporal features selected via their ANOVA F-Scores. Less of an erratic pattern is observed when compared to the statistical features. A relatively sharp increase of metrics can be observed towards the beginning, before becoming more gradual when 80 features have been introduced. Interestingly, this was also the number of features which stabilised the metrics for the statistical feature set. The best classifier when considering spatio-temporal features emerged when inputting 230, where a mean accuracy of 80.98\% was found. This model had a mean precision of 0.814, recall of 0.805, and F1-Score of 0.805.

\subsection{Early Fusion of Statistical and Spatio-temporal Features}
This section details the early fusion of both sets of features to combine them prior to making a prediction on the input data. 

\begin{figure}
    \centering
    \includegraphics[scale=0.75]{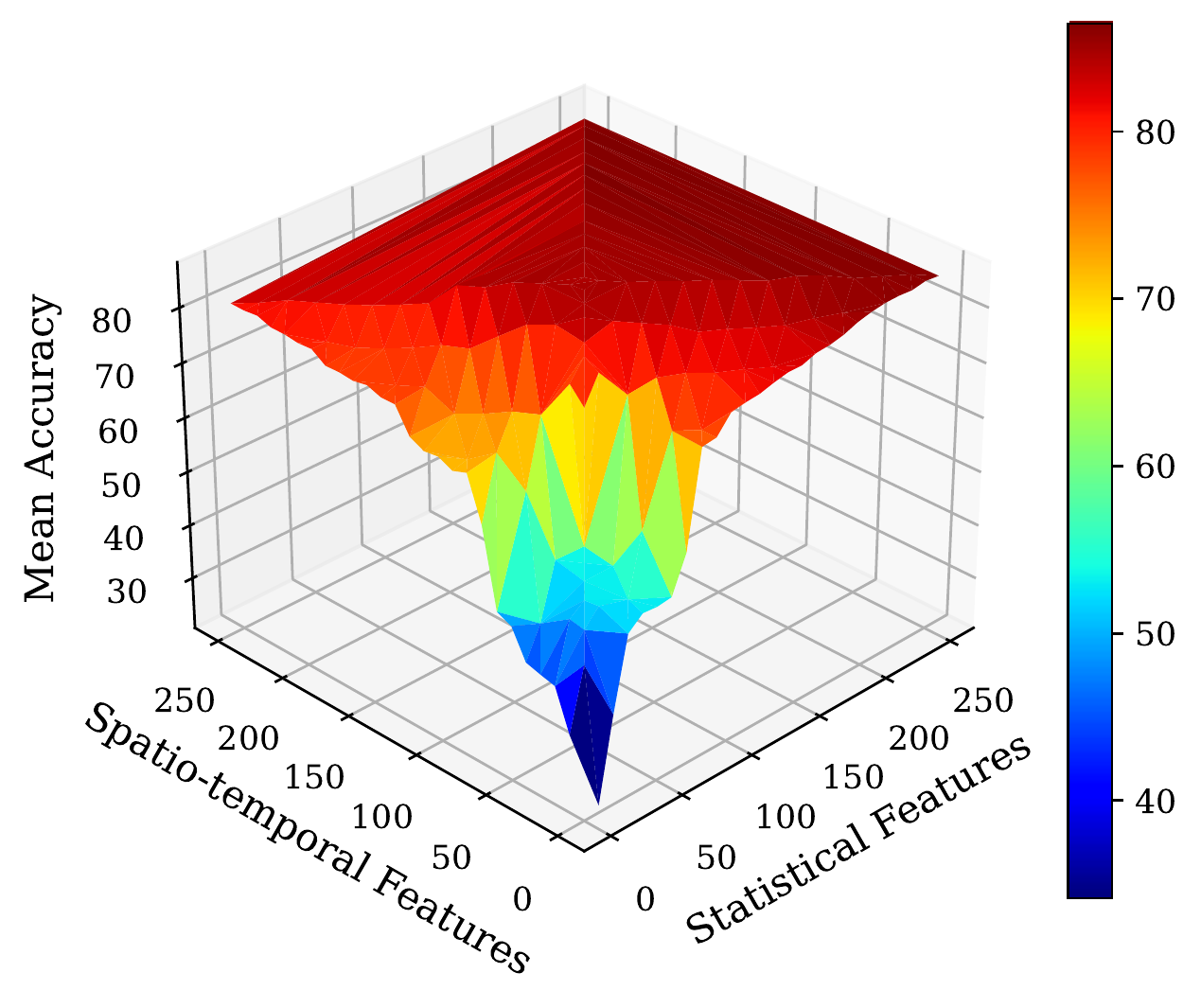}
    \caption{3D plot of the mean classification accuracy metrics when combining statistical and spatio-temporal features.}
    \label{fig:both-3d}
\end{figure}

\begin{figure}
    \centering
    \includegraphics[scale=0.75]{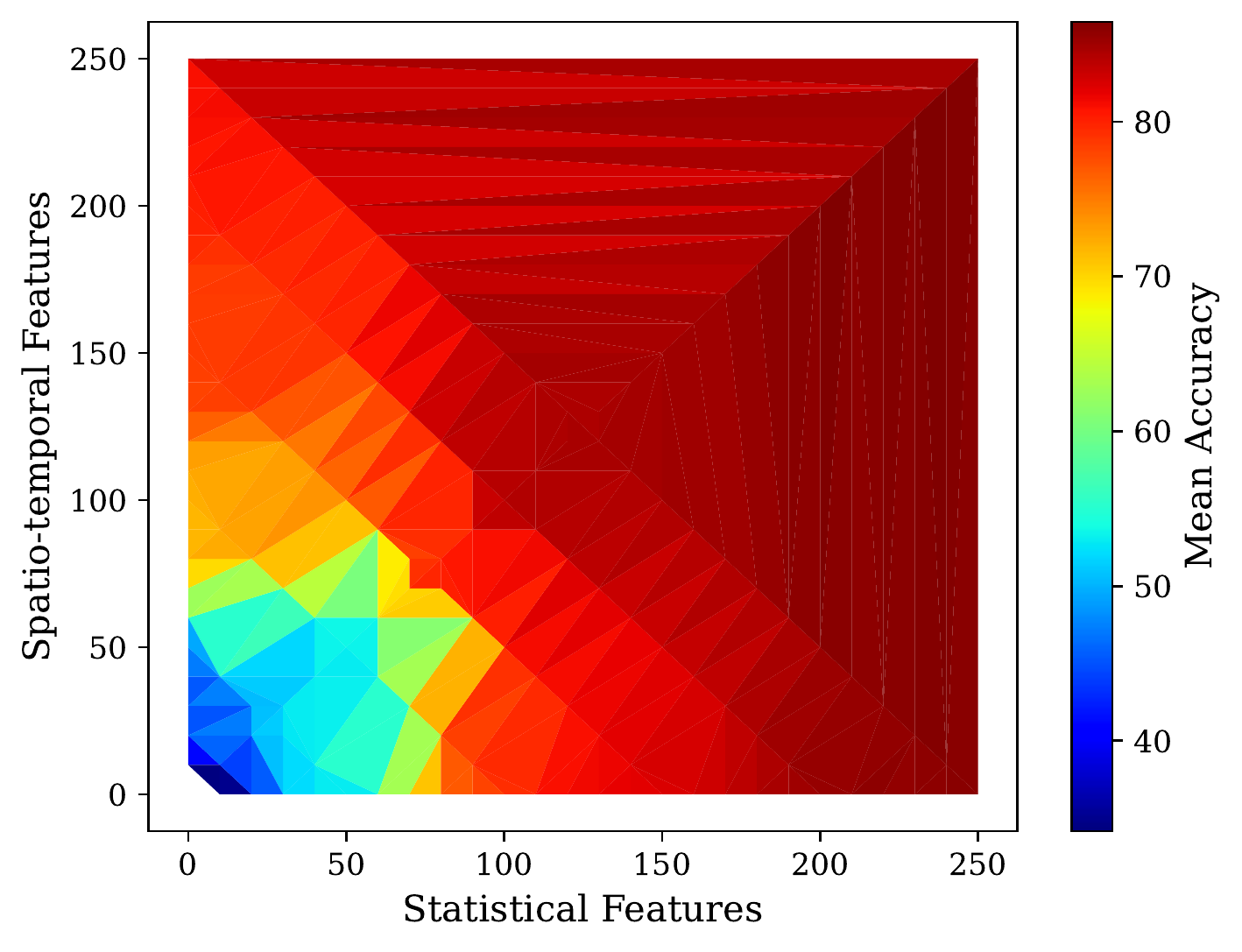} %
    \caption{Heatmap of the mean classification accuracy metrics when combining statistical and spatio-temporal features.}
    \label{fig:both-heatmap}
\end{figure}

Figures \ref{fig:both-3d} and \ref{fig:both-heatmap} show the a surface pertaining to the accuracy of the models when combining the sets of features. The single feature set results are also included in this surface (providing the two relevant edges). The points at which dataset dimensions are equal to or less than 250 are of higher resolution, since there are more combinations tested. The back half of the surface shows results when the selected number of features are equal for both sets. It can be observed that the highest values (darker shades of red) exist on the front half towards equal distribution and towards there being more statistical features, as well as for much of the shared feature selection surface.

\subsection{Comparison of Results}

\begin{table}[]
\centering
\caption{Ten best models observed from the set of all 146 machine learning experiments. Figures in parentheses show the K-Fold standard deviation.}
\label{tab:top-ten-results}
\footnotesize
\begin{tabular}{@{}rrrrrr@{}}
\toprule
\multicolumn{2}{l}{\textbf{Number of features}}                                                                                    & \multicolumn{4}{l}{\textbf{Mean K-Fold Metrics}}                                                                                                            \\ \midrule
\multicolumn{1}{l}{\textbf{Statistical}} & \multicolumn{1}{l}{\textbf{\begin{tabular}[c]{@{}l@{}}Spatio-\\ temporal\end{tabular}}} & \multicolumn{1}{l}{\textbf{Accuracy}} & \multicolumn{1}{l}{\textbf{Precision}} & \multicolumn{1}{l}{\textbf{Recall}} & \multicolumn{1}{l}{\textbf{F1}} \\
240                                      & 240                                                                                     & 86.75 (0.9)                                & 0.876 (0.085)                          & 0.864 (0.079)                       & 0.867 (0.066)                   \\
230                                      & 230                                                                                     & 86.66 (0.89)                               & 0.872 (0.087)                          & 0.863 (0.078)                       & 0.865 (0.067)                   \\
200                                      & 200                                                                                     & 86.6 (0.81)                                & 0.873 (0.086)                          & 0.86 (0.082)                        & 0.863 (0.068)                   \\
210                                      & 210                                                                                     & 86.48 (0.84)                               & 0.872 (0.086)                          & 0.861 (0.078)                       & 0.863 (0.067)                   \\
190                                      & 190                                                                                     & 86.36 (0.74)                               & 0.874 (0.094)                          & 0.859 (0.082)                       & 0.863 (0.072)                   \\
220                                      & 220                                                                                     & 86.3 (1)                                   & 0.872 (0.093)                          & 0.858 (0.075)                       & 0.862 (0.068)                   \\
180                                      & 180                                                                                     & 85.99 (1.04)                               & 0.868 (0.092)                          & 0.856 (0.085)                       & 0.858 (0.073)                   \\
240                                      & 0                                                                                       & 85.96 (0.51)                               & 0.866 (0.091)                          & 0.855 (0.081)                       & 0.857 (0.069)                   \\
240                                      & 10                                                                                      & 85.93 (0.83)                               & 0.867 (0.098)                          & 0.854 (0.081)                       & 0.857 (0.073)                   \\
250                                      & 250                                                                                     & 85.9 (0.83)                                & 0.868 (0.089)                          & 0.855 (0.078)                       & 0.858 (0.067)                   \\ \bottomrule
\end{tabular}
\end{table}
\begin{figure}
\centering
\begin{subfigure}{.5\textwidth}
  \centering
  \includegraphics[scale=0.5]{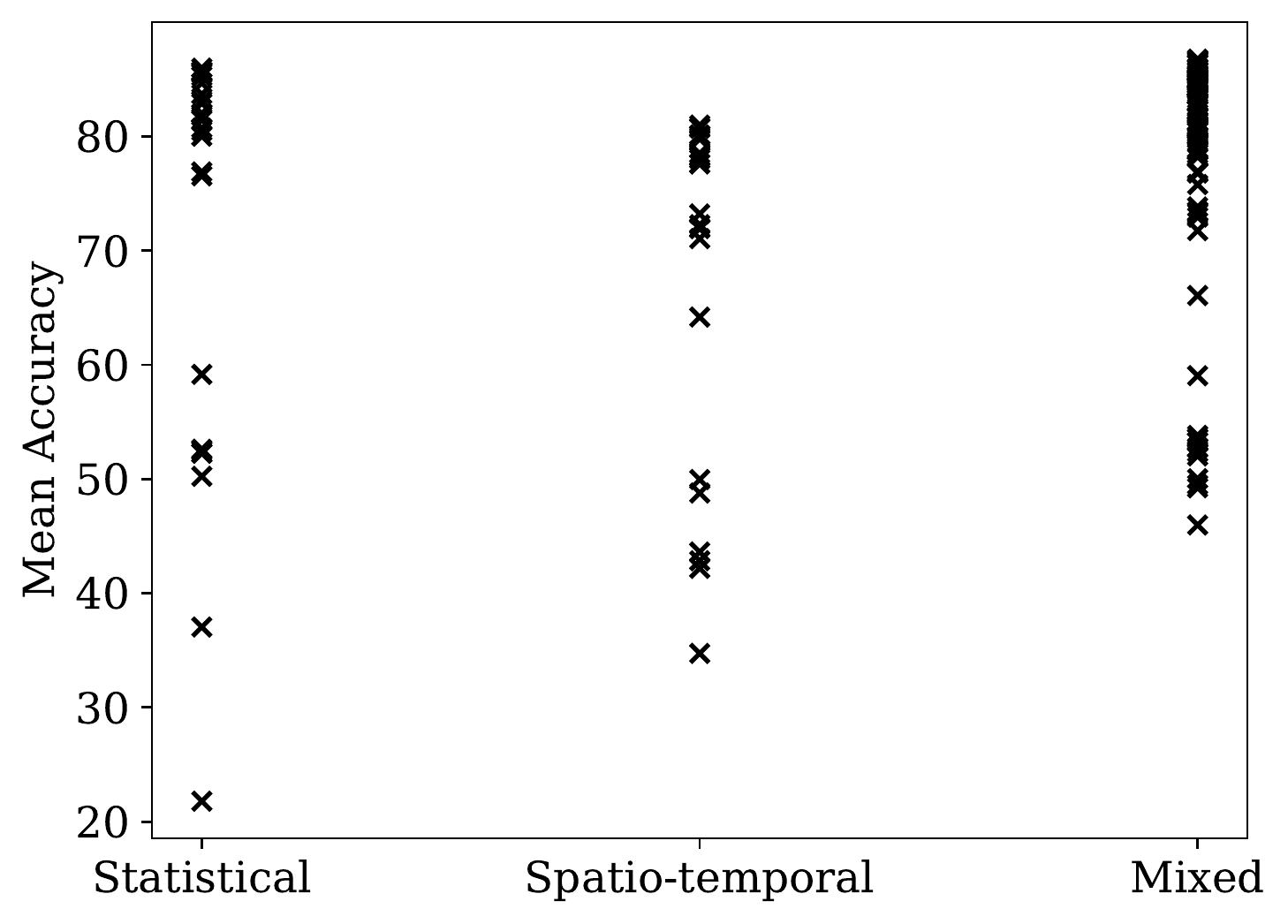}
  \caption{Scatter plot of all three sets of results.}
  \label{fig:sub1}
\end{subfigure}%
\begin{subfigure}{.5\textwidth}
  \centering
  \includegraphics[scale=0.5]{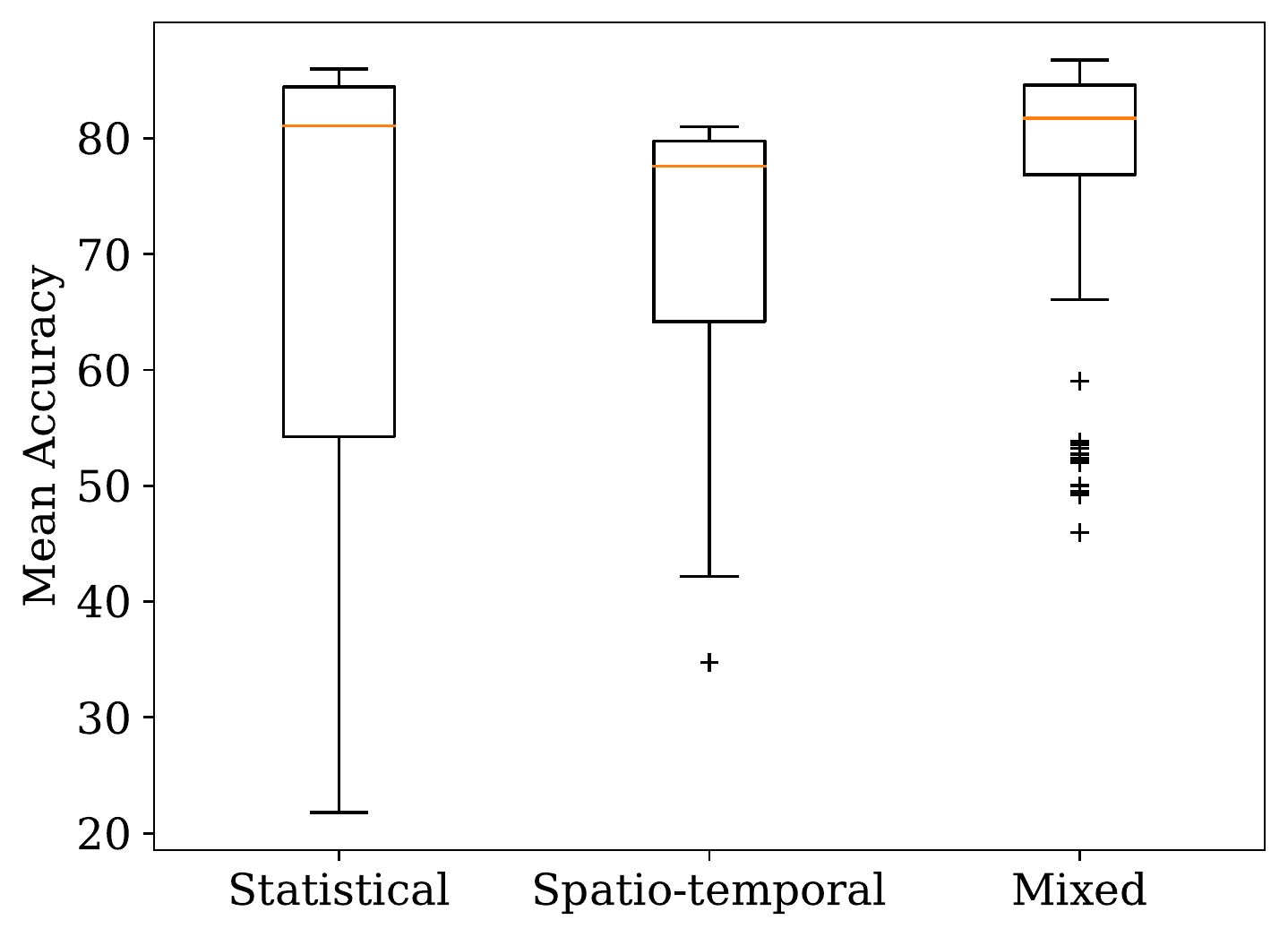}
  \caption{Box diagram of all three sets of results.}
  \label{fig:sub2}
\end{subfigure}
\caption{A comparison of how separate and mixed types of features affect the classification accuracy of the model.}
\label{fig:two-comps}
\end{figure}

The ten best models from all 146 machine learning experiments can be observed within Table \ref{tab:top-ten-results}. The best model overall combined the 240 best statistical and 240 best spatio-temporal attributes by their ANOVA F-Score. This model had a mean accuracy of 86.75\%, precision 0.876, recall 0.864, and F1-Score of 0.867. Note that the 8$^{th}$ best model was the first instance where a high score was encountered by using features from only one of the two domains. This is closely followed in 9$^{th}$ where only 10 spatio-temporal features are considered alongside 240 statistical features. Figure \ref{fig:two-comps} shows a scatter and box plot comparison of the three sets of features by their classification accuracy. Mixing features leads to a number of outliers towards the lower end of the plot, but the Q1, median and Q2 are observedly higher. No outliers were noted towards the maximum end of the results. Although the statistical features alone outperformed the spatio-temporal set in terms of the best results, it can be observed that the worst models outperformed those which were worst for the statistical set; this shows that when features are limited, considering temporal over statistical may lead to better results depending on how many the selection is limited to. Seven different combinations of both statistical and spatio-temporal features (which were equal in number) led to the strongest classification models in terms of their mean accuracy.
\begin{table}[]
\centering
\caption{Final best models observed when considering either one or both sets of features. Figures in parentheses show the K-Fold standard deviation.}
\label{tab:best-of-each}
\footnotesize
\begin{tabular}{@{}rrrrrr@{}}
\toprule
\multicolumn{2}{l}{\textbf{Number of features}}                                                                                    & \multicolumn{4}{l}{\textbf{Mean K-Fold Metrics}}                                                                                                            \\ \midrule
\multicolumn{1}{l}{\textbf{Statistical}} & \multicolumn{1}{l}{\textbf{\begin{tabular}[c]{@{}l@{}}Spatio-\\ temporal\end{tabular}}} & \multicolumn{1}{l}{\textbf{Accuracy}} & \multicolumn{1}{l}{\textbf{Precision}} & \multicolumn{1}{l}{\textbf{Recall}} & \multicolumn{1}{l}{\textbf{F1}} \\
240                                      & 240                                                                                     & 86.75 (0.9)                                & 0.876 (0.085)                          & 0.864 (0.079)                       & 0.867 (0.066)                   \\
240                                      & 0                                                                                       & 85.96 (0.51)                               & 0.866 (0.091)                          & 0.855 (0.081)                       & 0.857 (0.069)                   \\
0                                        & 230                                                                                     & 80.98 (0.69)                               & 0.814 (0.101)                          & 0.805 (0.097)                       & 0.805 (0.084)                   \\ \bottomrule
\end{tabular}
\end{table}
Table \ref{tab:best-of-each} then presents the final best models when selecting features from either both sets or just one set of features. Although better metrics are achieved, computational complexity must also be considered; at the cost of 0.79\% mean accuracy, the required number of features can be halved for a classification capability which is still relatively competitive. Stability is also negatively impacted as can be observed from the standard deviations of the scores when both attribute sets are present.

\section{Conclusion and Future Work}
To conclude, this study has explored how statistical and spatio-temporal feature extraction can be leveraged for the classification of sign language gestures. Results showed that mixing the two sets and learning through early fusion led to the best models overall in terms of their mean classification accuracy. When only single sets of features were considered, statistical features led to better classification over spatio-temporal, but the global minimum result was caused by extracting statistical features. It was also discovered that the worst results when mixing combinations of the features were higher than the worst models with only one type of feature as input. When the best ten out of all 146 trained models were compared by their classification metrics, the top 7 models were all combinations of mixed features, the eighth best was a model with statistical features only. Then, the ninth and tenth best models were also those which had a mixed set of features for learning.
The findings of this study have enabled much future work, firstly the number of chosen raw features prior to extraction was set based on an F-score cutoff point, future work could explore this figure to further improve the quality of the extracted features. On the point of feature selection, this study focused on F-scores for comparability, but other methods of selection could also be explored and compared to the results in this study. Finally, a random forest model was chosen due to its tendencies to generalise well and not overfit, and based on the results discovered through the trained 146 models, other methods of machine learning could be leveraged and compared.

\bibliographystyle{unsrt}  
\bibliography{references}

\end{document}